\documentclass[lettersize,journal]{IEEEtran}
\usepackage{amsmath,amssymb,amsfonts}
\usepackage{utfsym}
\usepackage{algorithmic}
\usepackage{algorithm}
\usepackage{array}
\usepackage[caption=false,font=normalsize,labelfont=sf,textfont=sf]{subfig}
\usepackage{textcomp}
\usepackage{stfloats}
\usepackage{url}
\usepackage{verbatim}
\usepackage{graphicx}
\usepackage{cite}
\usepackage{booktabs}
\usepackage{amsthm}
\usepackage{color}
\usepackage{xcolor}
\usepackage{soul}
\usepackage{multirow}
\usepackage{boldline} 
\usepackage{threeparttable}
\usepackage{hyperref}
\begin{document}

\title{Changes to Captions: An Attentive Network for Remote Sensing Change Captioning}

\author{Shizhen~Chang,~\IEEEmembership{Member,~IEEE,} and
        Pedram~Ghamisi,~\IEEEmembership{Senior Member,~IEEE}

\IEEEcompsocitemizethanks{\IEEEcompsocthanksitem S. Chang is with the Institute of Advanced Research in Artificial Intelligence, Landstraßer Hauptstraße 5, 1030 Vienna, Austria (e-mail: szchang@ieee.org; shizhen.chang@iarai.ac.at).
\IEEEcompsocthanksitem P. Ghamisi is with the Institute of Advanced Research in Artificial Intelligence, Landstraßer Hauptstraße 5, 1030 Vienna, Austria, and also with the Helmholtz-Zentrum Dresden-Rossendorf (HZDR), Helmholtz Institute Freiberg for Resource Technology (HIF), Machine Learning Group, Chemnitzer Str. 40, D-09599 Freiberg, Germany (e-mail: p.ghamisi@gmail.com; pedram.ghamisi@iarai.ac.at).}}

\markboth{IEEE Transactions on Image Processing, VOL. XX, NO. XX, August~2023}%
{Chang \MakeLowercase{\textit{et al.}}: A Sample Article Using IEEEtran.cls for IEEE Journals}


\maketitle

\begin{abstract}
In recent years, advanced research has focused on the direct learning and analysis of remote-sensing images using natural language processing (NLP) techniques. The ability to accurately describe changes occurring in multi-temporal remote sensing images is becoming increasingly important for geospatial understanding and land planning. Unlike natural image change captioning tasks, remote sensing change captioning aims to capture the most significant changes, irrespective of various influential factors such as illumination, seasonal effects, and complex land covers. In this study, we highlight the significance of accurately describing changes in remote sensing images and present a comparison of the change captioning task for natural and synthetic images and remote sensing images. To address the challenge of generating accurate captions, we propose an attentive changes-to-captions network, called Chg2Cap for short, for bi-temporal remote sensing images. The network comprises three main components: 1) a Siamese CNN-based feature extractor to collect high-level representations for each image pair; 2) an attentive encoder that includes a hierarchical self-attention block to locate change-related features and a residual block to generate the image embedding; and 3) a transformer-based caption generator to decode the relationship between the image embedding and the word embedding into a description. The proposed Chg2Cap network is evaluated on two representative remote sensing datasets, and a comprehensive experimental analysis is provided. The code and pre-trained models will be available online at https://github.com/ShizhenChang/Chg2Cap.
\end{abstract}

\begin{IEEEkeywords}
Change captioning, image captioning, remote sensing images, self-attention mechanism
\end{IEEEkeywords}

\section{Introduction}
\IEEEPARstart{C}{hange} captioning is an emerging research area at the intersection of natural language processing (NLP) and computer vision (CV), with the aim of providing meaningful descriptions of changes in a scene \cite{park2019robust}. This is particularly useful in enabling machine learning models to better understand the dynamic world we live in, similar to how humans process and interpret changes in their environment. With the increasing availability of multi-temporal remote sensing data, there has been a growing interest in using change captioning to study land cover changes and promote an intelligent interpretation of multi-temporal remote sensing images \cite{liu2022remote, hoxha2022change}. The ability to automatically generate accurate and detailed captions of land cover changes can have important applications in fields such as urban planning, disaster management, and environmental monitoring and can significantly improve our understanding of the changing nature of the built environment \cite{xu2022txt2img,wang2022glcm}.

Unlike traditional vision tasks, such as object detection \cite{chen2018scom, li2020harnet}, change detection \cite{chang2022deep, ru2020multi}, scene classification \cite{xu2020assessing}, and semantic segmentation \cite{xu2022consistency, xu2020assessing}, image captioning requires not only the identification of multiple objects in an image but also the generation of natural language sentences that accurately describe the relationships among those objects \cite{yang2020ensemble}. The change captioning task is particularly complex, as it involves interpreting image scenes from a sequential perspective and identifying the differences that occur over a period of time. In recent years, encoder-decoder-based architectures have emerged as popular solutions for change captioning. These architectures first extract deep features of the ``before" and ``after" scenes using Siamese neural networks, and then use decoders to translate these features into captions. Several methods have been proposed to improve the performance of change captioning models. Jhamtani et al. \cite{jhamtani2018learning} first introduced the change captioning task and proposed a Siamese CNN-RNN architecture to describe meaningful differences between similar images. Part et al. \cite{park2019robust} converted the semantic interpretation to a higher level and proposed a dual dynamic attention model that can distinguish semantic visual information between image pairs regardless of illumination and viewpoint distractions. To match the viewpoint distractors between the image pairs, Shi et al. \cite{shi2020finding} proposed a viewpoint-agnostic image encoder that exhaustively measures the feature similarity between image pairs. Yue et al. \cite{yue2023i3n} extended advancements by introducing an Intra- and Inter-representation Interaction Network (I3N) to finely learn different representations. In the Intra-representation Interaction stage, they devised a geometry-semantic interaction refining process to explore both positional and semantic interactions within the same image. In the Inter-representation Interaction stage, they employed hierarchical representation interaction to precisely identify latent differences in viewpoint changes. Furthermore, Tu et al. \cite{tu2023adaptive} introduced a viewpoint-adaptive representation disentanglement network that effectively discerns real changes from pseudo changes through position-embedded representation learning. To ensure the generation of reliable change features, they designed an unchanged representation disentanglement method.

Instead of proposing new model architectures, Hosseinzadeh et al. \cite{hosseinzadeh2021image} formulated a training scheme to improve the training accuracy of the existing change captioning network by means of an auxiliary task; an extra level of supervision was also provided through their training scheme. Guo et al. \cite{yao2022image} introduced a novel pretraining-finetuning paradigm for image difference captioning. To establish a robust association across modalities, they designed three self-supervised tasks with contrastive learning strategies. Building upon the impressive zero-shot performance of Contrastive Language-Image Pre-training (CLIP), they proposed an adaptation training process \cite{guo2022clip4idc} to tailor CLIP's visual encoder specifically for capturing and aligning image differences based on textual descriptions in the change captioning task. To enhance the adaptability of change captioning models to complex scenarios, Qiu et al. \cite{qiu2021describing} proposed a transformer-based network to describe more than one change between the image pair. The transformer-based architecture enhances the discernment of the network by generating new attention states for each token; thus, it performs well in both change captioning and localization.

As a leading-edge research topic in computer vision, change captioning has recently gained attention in the geoscience and remote sensing field due to its ability to extract high-level semantic information about changes in land cover. Compared to direct change detection methods, captioning is more flexible and can describe different types of changes. To encourage research in this area, several related works and datasets have been proposed. Hoxha et al. \cite{hoxha2022change, chouaf2021captioning} first proposed image-based level and feature-based level fusion strategies to concatenate bitemporal images as input to the caption generator and published two small datasets to address the absence of remote sensing change captioning (RSCC). To further advance research in this field, Liu et al. \cite{liu2022remote, liu2023progressive} then created a large RSCC dataset, called LEVIR-CC, and used dual transformers and a vision transformer (ViT) as part of the encoder, with a transformer-based decoder to generate captions.

\begin{figure*}
    \centering
    \includegraphics[width=0.95\linewidth]{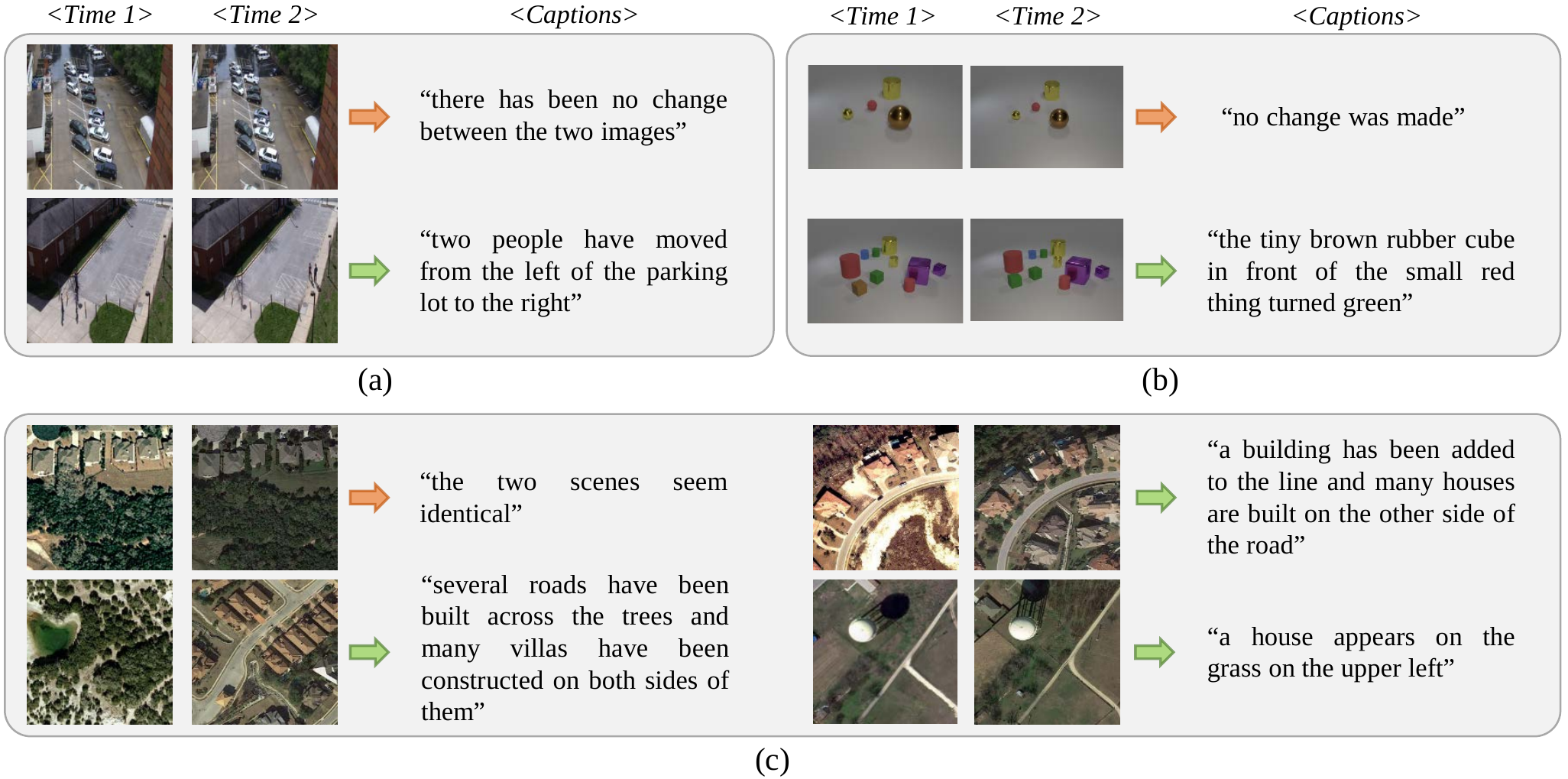}
    \caption{Examples of change captioning in (a) Spot-the-Diff dataset. (b) CLEVR-Change dataset. and (c) LEVIR-CC dataset. Spot-the-Diff is a natural image dataset captured from video surveillance, CLEVR-Change is a synthetic dataset that simulates viewpoint and illumination distractions happening between image pairs, and LEVIR-CC is a remote sensing dataset. The \textcolor{orange}{Orange} arrows connect image pairs and descriptions for unchanged scenes, and the \textcolor{lime}{Green} arrows are for changed scenes.}
    \label{fig1}
\end{figure*}

These pioneering publications have set the foundation for the development of change captioning in remote sensing. Unlike natural images, remote sensing images present unique opportunities and challenges for change captioning. Here, we elucidate the significance of pursuing the remote sensing change captioning task:

\begin{itemize}
    \item Remote sensing image pairs are visually different due to variations in illumination intensity, direction, and seasonal effects. This makes it necessary to account for distribution gaps between original features of the image pairs caused by acquisition conditions when designing efficient change captioning networks for remote sensing images. For example, the time interval between two remote sensing images can be months or years, unlike natural images, which are usually collected in a short period of time. Fig. \ref{fig1} provides examples of change captioning scenes from natural and remote sensing images to illustrate this point.
    \item Captions of changed scenes are more complex in remote sensing data. Compared to natural images, remote sensing images require more robust and accurate descriptions of changes. While natural image frame pairs extracted from videos are mostly well aligned, and the network focuses on describing the movements of people and vehicles, the synthesized datasets generated by the image generation engine introduce viewpoint and lighting distractions to simulate multiple compositional scenes, as shown in Fig. \ref{fig1}(a) and (b). However, the caption generation network for remote sensing images must generate accurate descriptions of changes regardless of the complex distribution of land covers.
    \item Objects in remote sensing images are more difficult to distinguish than in natural images. Since the remote sensing images are taken at high altitudes vertically, most of the three-dimensional information, such as color, height, and texture, is compressed into two-dimensional information of the Earth's surface in remote sensing images. This makes it more difficult to identify the species of object from a top-down perspective. Therefore, change captioning networks need to be sensitive to changed land covers without height and other details.

    \item Change captioning can yield a better understanding of urban planning. Remote sensing techniques are closely related to geographic interpretation, and change captioning networks can help researchers augment geo-information for the region of interest with appropriate textual descriptions and semantic features. Captions focusing on changes can help people intuitively visualize which part of the scene has been removed or developed, saving urban planners the effort of assessing and analyzing. Additionally, RSCC networks require descriptions suitable for geographic changes of interest while disregarding irrelevant interference.
\end{itemize}

To generate precise change descriptions for remote sensing bitemporal images, we propose a novel changes-to-captions network, called Chg2Cap, based on attention mechanisms. The Chg2Cap architecture comprises an encoder-decoder framework, featuring a hierarchical self-attentive block and a residual block integrated into the attentive encoder. By stacking self-attention mechanisms, the network gains the ability to hierarchically capture inter- and intra-information of the deep features. The residual block, enhanced with the cosine mask, contributes to enhancing the consistency and inconsistency between the retrieved feature pairs. Additionally, we improve the transformer-based caption generator with residual connections to preserve the local information and progressively decode feature embeddings.

Our work contributes in three main ways:
\begin{enumerate}
\item We emphasize the importance and necessity of designing an appropriate change captioning method for remote sensing images, considering their unique nature compared to regular natural images. Through representative examples selected from both natural and synthetic image datasets, as well as remote sensing datasets, we highlight the distinctive characteristics of remote sensing image pairs. Moreover, we underscore the potential significance of applying change captioning in remote sensing for a better understanding of geo-information.
\item We propose the Chg2Cap method, which utilizes an attentive encoder and a transformer-based decoder for remote sensing change captioning. The attentive encoder captures the inter- and intra-information through the hierarchical self-attention block and enhances the consistency and inconsistency features using the residual block. Through extensive experiments, we demonstrate that the proposed Chg2Cap can generate more accurate and realistic descriptions of the changes and achieves the best performance when compared to existing change captioning methods.
\item We comprehensively compare and analyze the effects of attention mechanisms in both image feature representation and caption generation stages. By conducting systematic parametric analysis and evaluating different network settings, we provide insights that may inspire researchers to design more appropriate models and make full use of bitemporal features.
\end{enumerate}

The rest of this paper is organized as follows. Section II reviews works related to this study. Section III describes the proposed Chg2Cap method in detail. Section IV presents the information on datasets used in this study and the experimental results. Conclusions and other discussions are summarized in Section V.

\section{Related Works}
In this section, we will introduce the prior research on image-to-caption generation and attention mechanisms in image captioning.
\subsection{Image-to-Caption Generation}
As an interdisciplinary field at the intersection of CV and NLP, image captioning \cite{stefanini2022show, huang2020image, yuan2022discriminative} has gained significant attention in recent years, with many researchers applying advanced artificial intelligence techniques to improve the state of the art. In this section, we provide a brief review of image captioning in both computer vision and remote sensing fields.

There are generally three categories of image captioning methods based on how the descriptions are generated: template-based methods, retrieval-based methods, and sequence generation-based methods. Template-based methods rely on pre-defined sentence structures with blank slots to fill in object names and attributes to generate descriptions. Retrieval-based methods, on the other hand, use a large-scale database of images and corresponding descriptions to find the most similar image to the input and retrieve its description as the output. While these methods can generate grammatically correct sentences, they are limited by the fixed templates and database of information. With the advent of deep learning and NLP techniques, sequence generation-based methods have been developed which treat captions as sequences of word tokens and train a model to learn the relationship between image embeddings and word embeddings. These methods have been shown to be more flexible and capable of generating more diverse and apt descriptions.

Sequence generation-based methods, which follow the encoder-decoder paradigm of neural machine translation, have been widely used to generate more flexible descriptions of input images. In recent years, this approach has also been applied to remote sensing image captioning (RSIC). Qu et al. \cite{qu2016deep} were among the first to propose using the CNN-RNN architecture to extract features and generate captions for remote sensing images. They also published the first two datasets for RSIC, namely Sydney-captions and UCM-captions. Later, Lu et al. \cite{lu2017exploring} proposed a larger dataset for remote sensing image captioning, taking into account different people's interpretations of an image and describing various sentence patterns in a more appropriate format. They evaluated a variety of baseline models borrowed from the natural image captioning task and discussed the transferability of different captioning networks from natural to remote sensing images. It is worth mentioning that extensive experimental evaluations have highlighted the need for more methods tailored to the characteristics of remote sensing images for remote sensing datasets.

Recently, Hoxha et al. were the first to propose a framework for generating change captions in multi-temporal remote sensing images, which has been evaluated using different fusion strategies on two small datasets they published \cite{hoxha2022change, chouaf2021captioning}. To further advance research in this emerging field, Liu et al. created a large-scale RSCC dataset by augmenting their previous building change detection dataset \cite{liu2022remote, liu2023progressive}. They employed transformer-based architectures to generate descriptions for the image pairs, which demonstrated improved performance over previous studies.

\subsection{Attention in Image Captioning}
To improve the realism of generated descriptions, attention mechanisms have been incorporated into the encoder-decoder paradigm, which is widely used in image captioning tasks \cite{huang2019attention, ji2020spatio, fu2016aligning}. Xu et al. \cite{xu2015show} were among the first to propose using attention mechanisms to extract important information for the task at hand. They introduced soft and hard attention mechanisms and visualized attention weights to provide insight into the model's decision-making process. Lu et al. \cite{lu2017knowing} further developed an adaptive attention mechanism that reweights image and text features based on their relevance to the task. This adaptive encoder-decoder framework can automatically determine when and where to ``look" when generating words, resulting in more human-like captions and question answers. Anderson et al. \cite{anderson2018bottom} combined bottom-up and top-down attention mechanisms to enable attention to be calculated at the level of objects and other salient image regions, which improved the accuracy of object recognition and localization. In the domain of remote sensing image captioning, several notable methods have been introduced by different researchers. Zhao et al. \cite{zhao2021high} presented a fine-grained, structured attention-based approach that leverages the structural characteristics of semantic contents to generate descriptive captions. Zhuang et al. \cite{zhuang2021improving} devised a transformer encoder-decoder architecture, effectively incorporating grid visual features to produce accurate and precise captions. Liu et al. \cite{liu2022remote2} adopted a multi-layer aggregated transformer to comprehensively understand input information from different levels, resulting in improved caption generation.

To improve the network's focus on changes, recent studies have proposed novel attention-based models for change captioning. Park et al. \cite{park2019robust} developed a dual dynamic attention model that utilizes spatial dual attention to identify changes and semantic attention to highlight differences in visual representations. Tu et al. \cite{tu2023neighborhood} designs a common feature distilling module to compare the features and learn an effective contrastive representation. More recently, Liu et al. \cite{liu2022remote} adopted the transformer architecture as the caption generator for RSCC. Unlike traditional recurrent neural networks (RNNs), the transformer-based architecture can model intra- and inter-modality visual relationships effectively and generate precise descriptions.

In this paper, we propose the use of attention mechanisms in both the image embedding extraction and caption generation stages to accurately locate changes between bitemporal image features and describe the changes that have occurred in the corresponding scenes. A detailed discussion of the proposed attentive network is provided in Section III.
\section{Methodology}
\begin{figure*}
    \centering
    \includegraphics[width=\linewidth]{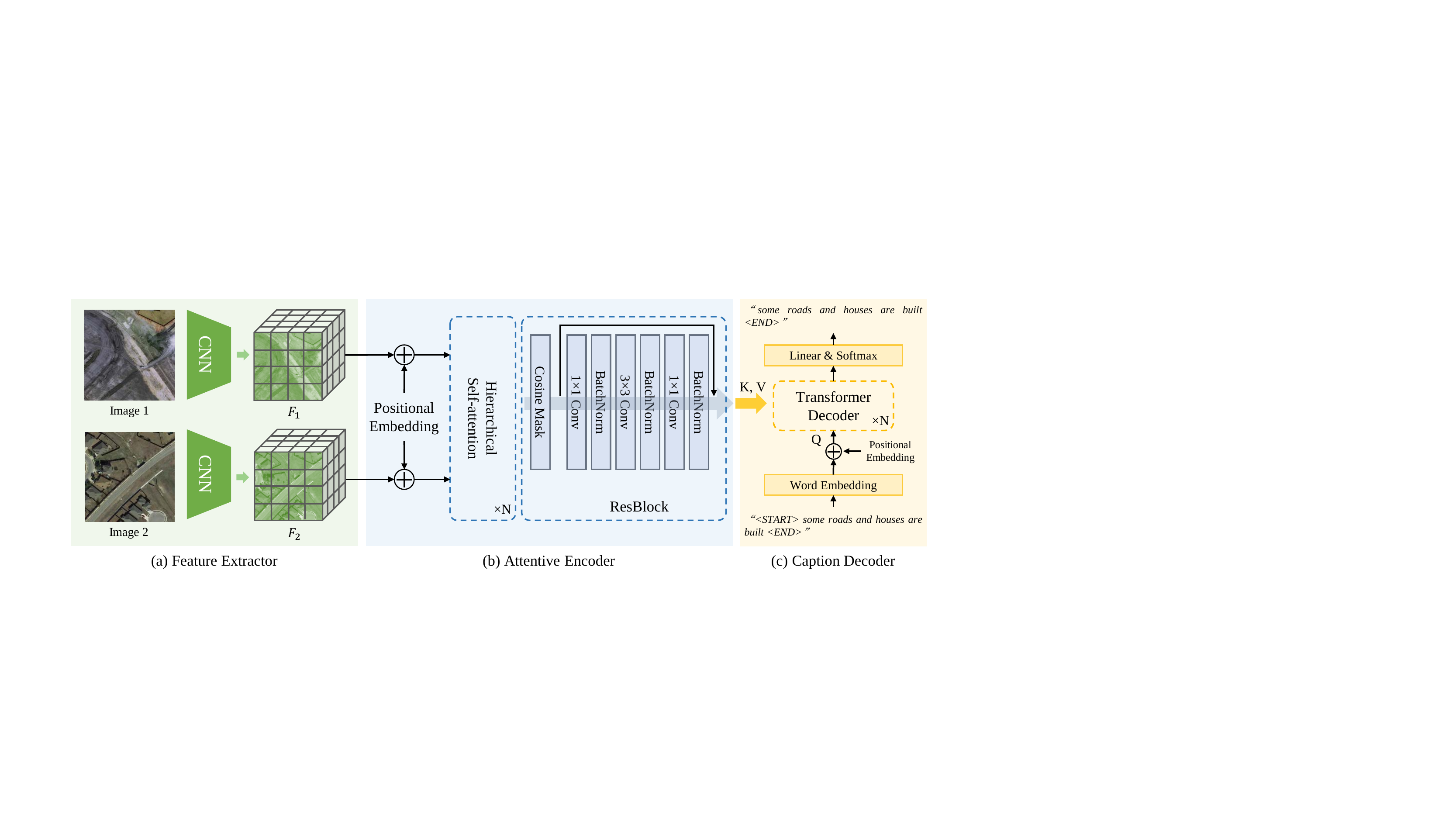}
    \caption{Overall framework of the proposed Chg2Cap method which is constructed by (a) a CNN-based feature extractor, (b) an attentive encoder consisting of N stacks of hierarchical self-attention (HSA) blocks and a residual block (ResBlock), and (c) a caption generator.}
    \label{fig2}
\end{figure*}
\subsection{Overview}
The proposed Chg2Cap framework consists of a CNN-based feature extractor, an attentive encoder, and a caption decoder, as shown in the flowchart in Fig. \ref{fig2}. In the training phase, the network takes a bitemporal remote sensing image pair with corresponding change description as input. First, the CNN backbone extracts deep features from the input images. Next, the bitemporal deep features are fed into the attentive encoder, where a hierarchical self-attention block followed by a residual block (ResBlock) is designed to localize change information within the image features. Finally, the decoder utilizes both the image embeddings and the word embeddings to generate the predicted change captions by capturing the inter-relationship between them. Formally, we denote a pair of training images collected from time 1 and time 2 as $X = (X_1, X_2)$, respectively, and the corresponding human-annotated text tokens that describe the changes between the image pairs as $t = (t_1; t_2; ...; t_n)$, where $n$ is the maximum length of the sentence. The text tokens start with the ``START" token and end with the ``END" token. If the number of text tokens is less than $n$, we pad $0$ as placeholders to fill the empty tokens. Chg2Cap acts as a function $f(X, t; \theta) = \hat{t}$ parameterized by $\theta$ that takes $X$ and $t$ as inputs, and outputs the predicted change captions. The whole network is optimized by minimizing the cross-entropy loss between the probability vector of the predicted change captions and the original text tokens.

In the validation and test phases, due to the absence of textual descriptions, the predicted captions are generated autoregressively based solely on the input image pairs. In the context of image captioning, autoregression involves generating a caption one word at a time, where each predicted word is conditioned on the previously generated words. Specifically, the caption generation process starts with the initialization of the ``START" token, and subsequent predictions are conditioned on previous predictions.
\subsection{Feature Extractor}
In the context of analyzing pairs of images $(X_1, X_2)$, a common approach is to use a Siamese Convolutional Neural Network (CNN) with shared weights as a feature extractor. In this paper, we adopt the widely-used ResNet101 architecture \cite{he2016deep}, which has been pre-trained on the ImageNet dataset, as the backbone of our network. Specifically, the input images are fed through the ResNet101 network, omitting the dropout layer. The deep feature pairs $(F_1, F_2)$ that represent the underlying content of the images are obtained as follows:
\begin{equation}
\begin{split}
    F_1 &= f_{fe}(X_1;\theta_{fe}); \\
    F_2 &= f_{fe}(X_2;\theta_{fe}),
\end{split}
\end{equation}
where $f_{fe}$ and $\theta_{fe}$ denote the function and parameters of the feature extractor, respectively. By utilizing a Siamese architecture and sharing weights between the two networks, we can compare the features extracted from two images efficiently and with reliability. As a result, we are able to identify and analyze any discrepancies between the two images during the subsequent stages of the proposed methodology.

\subsection{Attentive Encoder}
\begin{figure}
    \centering
    \includegraphics[width=0.95\linewidth]{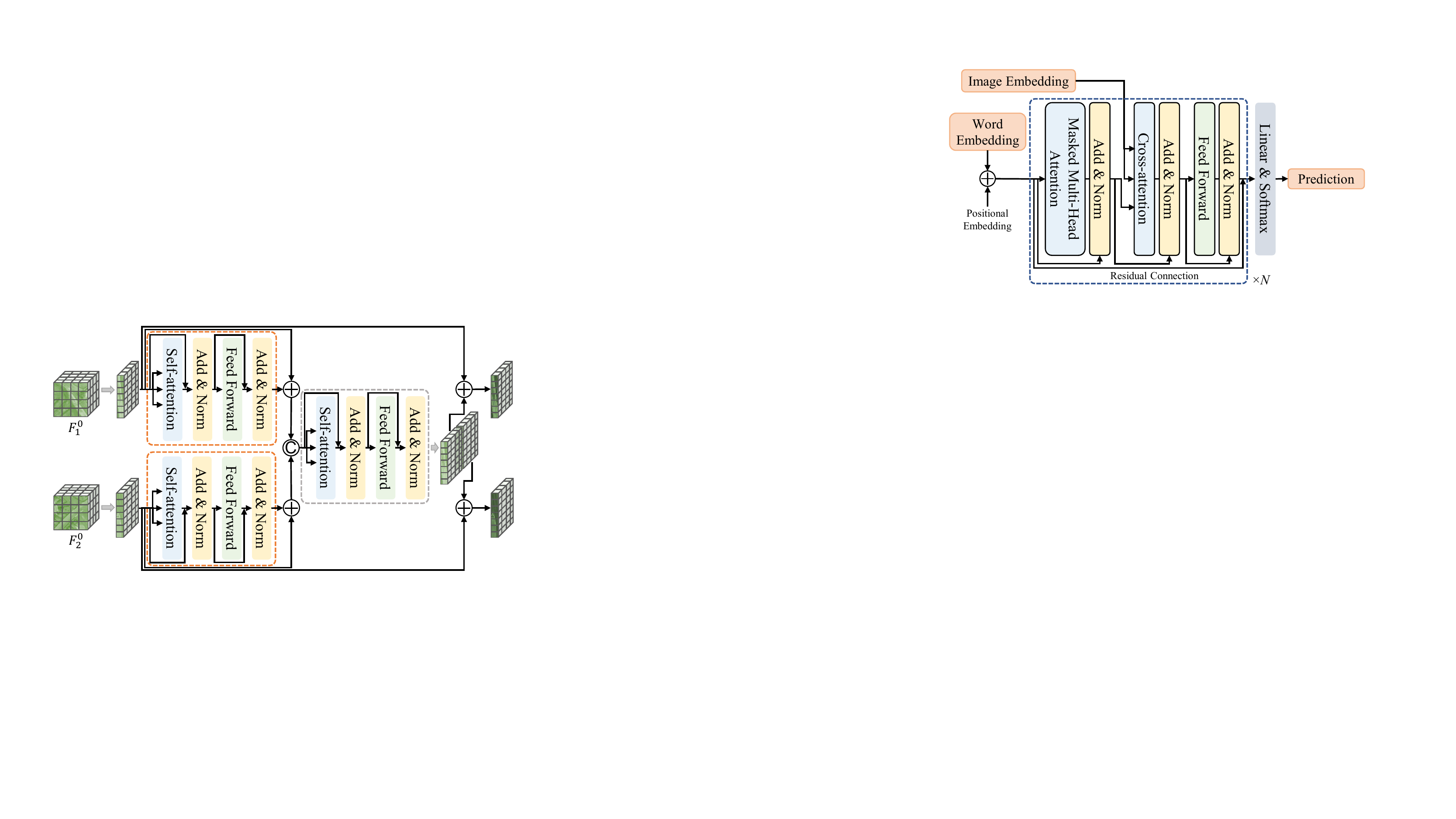}
    \caption{Visualization of the hierarchical self-attention (HSA) block. The deep features of each image are initially passed through a dual self-attention unit (framed by \textcolor{orange}{orange} dashed lines) with shared attention weights. The resulting features are then concatenated with residual connections before being passed to the joint self-attention unit (framed by \textcolor{gray}{gray} dashed lines). Finally, the hierarchically self-retrieved feature pairs are obtained.}
    \label{fig3}
\end{figure}

The attentive encoder that employs the self-attention mechanisms is designed to localize changes between deep feature pairs. The encoder consists of two blocks: a stack of hierarchical self-attention (HSA) blocks and a residual block (ResBlock) that fuses the bitemporal features into an image embedding. This approach enables the attentive encoder to capture subtle and complex changes between deep feature pairs, which may be challenging to identify using traditional image analysis techniques. As shown in Fig. \ref{fig3}, the HSA block captures both inter and intra-representations using a dual self-attention (DSA) unit and a joint self-attention (JSA) unit. The DSA unit computes the semantic information of individual deep features, while the JSA unit performs attention on concatenated features. 

The HSA block first takes in features with positional embeddings as initial inputs:

\begin{equation}
    F_i^0 = F_i + F_{pos}, \ \ \ i= 1, 2,
\end{equation}
where $i$ represents the temporal dimension of the image pairs and $F_{pos}\in\mathbb{R}^{h\times w\times C}$ is a learnable 2D position embedding that incorporates the spatial position information into bitemporal feature sequences \cite{qiu2021describing}. 

The features $F_1^0$ and $F_2^0$ are then transformed into two-dimensional matrices with sizes $hw\times C$. The processing of the $j$-th HSA block can be expressed as
\begin{equation}
    [Q^{j-1}_i; K^{j-1}_i; V^{j-1}_i] = F^{j-1}_i W, \ \ \ W\in\mathbb{R}^{C\times 3C_t},
\end{equation}
\begin{equation}
    F^j_i=\text{DSA}^j(Q^{j-1}_i, K^{j-1}_i, V^{j-1}_i)+F^{j-1}_i,
\end{equation}
\begin{equation}
    F^j_{cat}=[F^{j}_1; F^{j}_2],
\end{equation}
\begin{equation}
    [Q^{j}_{cat}; K^{j}_{cat}; V^{j}_{cat}] = F^{j}_{cat} W_{cat}, \ \ \ W_{cat}\in\mathbb{R}^{2C\times 3C_t},
\end{equation}
\begin{equation}
    [\Tilde{F}^{j}_1; \Tilde{F}^{j}_2] = \text{JSA}^j(Q^{j}_{cat}, K^{j}_{cat},V^{j}_{cat}),
\end{equation}
\begin{equation}
    F^j_i = \Tilde{F}^{j}_i + F_i^{j-1},
\end{equation}
where $j=1,...,N$ denotes the depths of the HSA block, $[\ ;\ ]$ denotes the concatenation operation, and $W$ and $W_{cat}$ are the weight matrices that linearly map the input to the query $Q$, key $K$, and value $V$, simultaneously. In this work, the dimension of the projection weights $C_t$ is set equal to $C$. $\text{DSA}^j$ and $\text{JSA}^j$ denote the function of $j$-th dual self-attention unit and joint self-attention unit, respectively. As shown in Fig. \ref{fig3}, the architecture of these two units is consistent. The input attends to retrieve a weighted sum of the embedding through the self-attention layer, and the retrieved feature is passed to a fully connected feed-forward network. To allow the gradients of the initial input feature to flow directly to the output and enable the model to preserve important information, we apply two residual connections after the DSA unit and the JSA unit.

In the ResBlock, we aim to construct accurate image embedding for the caption decoder. To further enhance the consistency and inconsistency information between the feature pairs $(F_1^N, F_2^N)$, we calculate the cosine similarity between the two features and add it to the concatenated features. The resulting feature map is then passed through a residual block consisting of three 2D convolutional layers, each followed by a Rectified Linear Unit (ReLU) activation function. Mathematically, the computation performed by the ResBlock can be expressed as:

\begin{equation}
    Mask = \text{Cos}(F_1^N, F_2^N),
\end{equation}
\begin{equation}
    F_{fus} = [F^N_1; F^N_2] + Mask,
\end{equation}
\begin{equation}
\begin{split}
    &\Tilde{F}_{fus}= Conv_3(ReLU(Conv_2(ReLU(Conv_1(F_{fus}))))),
\end{split}
\end{equation}
\begin{equation}
    E_{img}= LN(\Tilde{F}_{fus}+F_{fus}),
\end{equation}
where $\text{Cos}(\cdot, \cdot)$ calculates the cosine similarity between the two matrices, resulting in a $hw\times 1$ vector. $F_i^N(i= 1, 2)$ are the retrieved features obtained from the previous HSA block. The fused feature $F_{fus}$ is then derived by element-wise addition between the concatenation of $F^N_1$ and $F^N_2$ and the broadcasted $Mask$. $Conv_1$, $Conv_2$, and $Conv_3$ are three 2D convolutional layers with kernel sizes of $1\times 1$, $3 \times 3$, and $1 \times 1$, respectively. $LN$ represents the layer normalization. The output image embedding $E_{img}$ in the shape of $hw\times d_{emb}$ is finally derived. 
\subsection{Caption Decoder}
\begin{figure}
    \centering
    \includegraphics[width=0.95\linewidth]{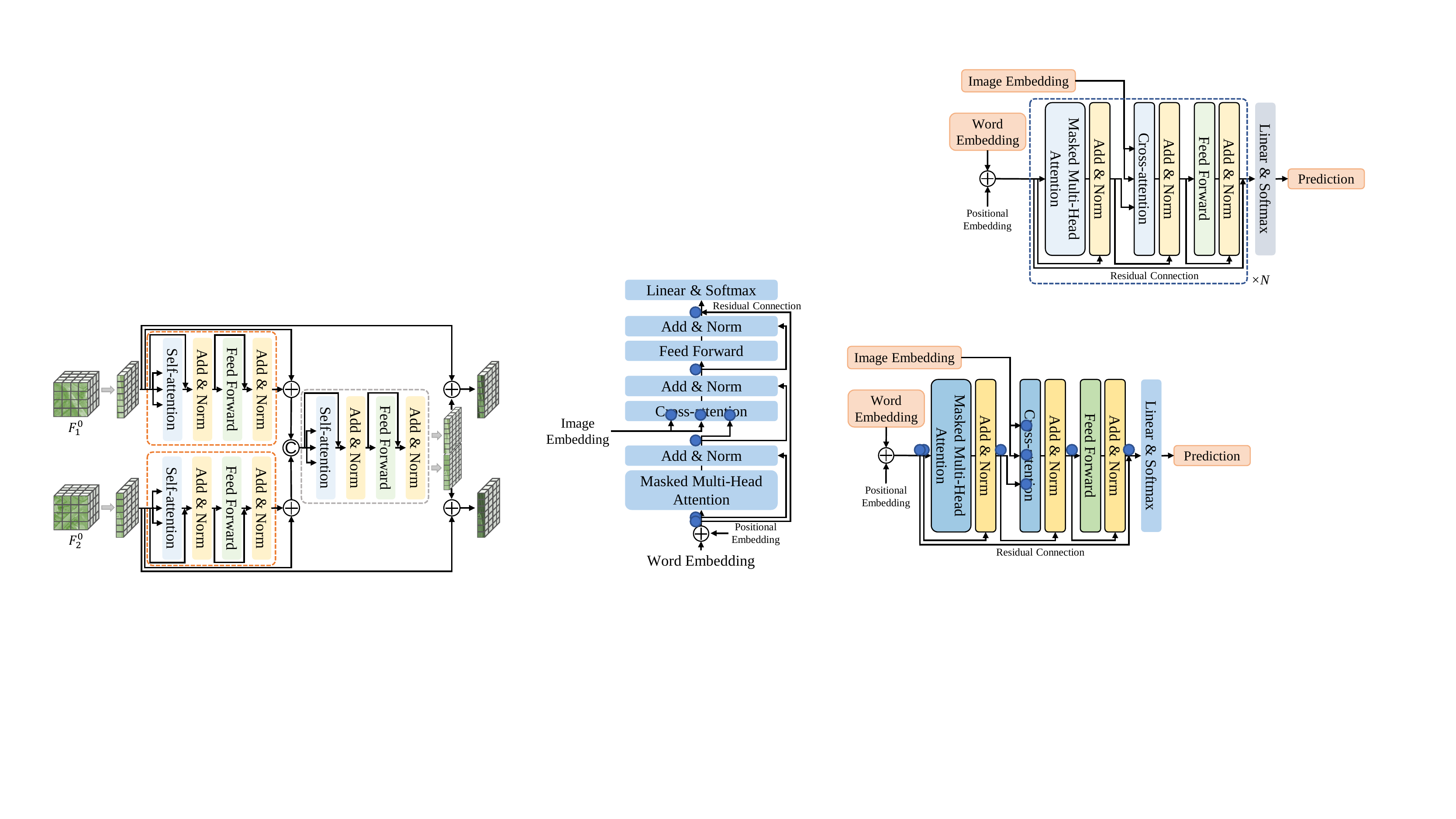}
    \caption{Visualization of the proposed caption generator. To preserve the information of input tokens locally, a residual connection of the word embedding is introduced into the original transformer-based decoder.}
    \label{fig4}
\end{figure}

To accurately capture and describe the changes between pairs of images, we propose to leverage the transformer-based decoder, which has been widely adopted in various natural language processing tasks recently, for our Chg2Cap method. The decoder consists of multiple transformer decoder layers, each of which comprises a masked multi-head attention sub-layer, a cross-attention sub-layer that integrates the encoder-decoder embeddings, and a feed-forward network. To maintain information flow and ensure model stability, these sub-layers are followed by a residual connection and a layer normalization operation. Moreover, to preserve the information of the word embedding, we employ a residual connection for the entire decoder layer. Finally, the output embedding is generated via a linear layer and a softmax activation function. Fig. \ref{fig4} provides a visual representation of this structure.

To prepare the change descriptions for the caption decoder during training, the text tokens are first mapped to a word embedding using an embedding layer. Specifically, we use a function $f_{emb}: \mathbb{R}^n \rightarrow \mathbb{R}^{n\times d_{emb}}$ to map the original tokens $t$ to the word embedding $E_{text}$. The initial inputs of the transformer decoder can then be obtained by:
\begin{equation}
    E^0_{text}=f_{emb}(t)+E_{pos},
\end{equation}
where $E_{pos}$ is the positional embedding that is calculated based on the sinusoidal function with distinct frequencies and phases for each token's position, as described in \cite{vaswani2017attention,devlin2018bert}.
Then the $j$-th masked multi-head attention sub-layer in a stack of $N$ layers can be formulated as

\begin{equation}
\begin{split}
S^j_{text} &= \text{MHA}^j(E^{j-1}_{text},E^{j-1}_{text}, E^{j-1}_{text}) \\
&=\text{Concat}(head_1, ..., head_h)W^O,
\end{split}
\end{equation}
where
\begin{equation*}
\begin{split}
&head_l= \text{Attention}(E^{j-1}_{text}W_l^Q,E^{j-1}_{text}W_l^K,E^{j-1}_{text}W_l^V),  \\
&\text{Attention}(Q,K,V) = \text{softmax}(\frac{QK^\top}{\sqrt{d_K}})V.
\end{split}
\end{equation*}
$\text{MHA}^j$ denotes the function of the $j$-th masked multi-head attention sub-layer, $h$ is the number of heads in the masked multi-head attention sub-layer, $W_l^Q, W_l^K, W_l^V\in\mathbb{R}^{d_{emb}\times \frac{d_{emb}}{h}}$ are trainable weight matrices of the $l$-th head, and $W^O\in\mathbb{R}^{d_{emb}\times d_{emb}}$ is a trainable weight matrix that performs linear projection to the output feature dimension. It is important to note that attention mechanisms can cause the information to ``leak" to future positions in the output sequence, which may compromise the accuracy of the model. To prevent this, a lower-triangular mask with the same shape as the attention weights matrix is created. This mask ensures that the model only attends to positions that have already been generated in the output sequence, thereby avoiding any leakage of information.

Similarly, the cross-attention sub-layer in the decoder also employs a multi-head attention mechanism, where the query comes from the previous masked multi-head attention sub-layer and the key and value come from the attentive encoder. Mathematically, the output of the $j$-th cross-attention sub-layer can be expressed as
\begin{equation}
S^j_{text-img}= \text{CA}^j(S^j_{text},E_{img},E_{img}),
\end{equation}
where $\text{CA}^j$ represents the function of the $j$-th cross-attention sub-layer.

The feature $S^j_{text-img}$ is subsequently processed by a feed-forward network, and the final output of the transformer decoder layer is obtained by adding the input word embedding with the output of the feed-forward network through a residual connection:
\begin{equation}
E^{j}_{text}=\text{FN}^j(S^j_{text-img})+E^{j-1}_{text},
\end{equation}
where $\text{FN}^j$ denotes the function of the feed-forward network.

After $N$ layers of transformer decoder, the retrieved word embedding is fed to a linear layer followed by a softmax activation function to obtain the word probabilities, denoted by $T=[\hat{\boldsymbol{t}}_1;\hat{\boldsymbol{t}}_2;...;\hat{\boldsymbol{t}}_n]\in\mathbb{R}^{n\times m}$, where $n$ is the length of the generated caption and $m$ is the size of the vocabulary. The model is trained by minimizing the cross-entropy loss between the predicted word probabilities and the reference captions. The loss function is defined as:

\begin{equation}
\mathcal{L}=-\sum_{i=1}^{n}\sum_{j=1}^{m}t_{i,j}\log \hat{t}_{i,j},
\end{equation}
where $t_{i,j}$ is the one-hot encoding of the $i$-th word in the reference caption at time step $j$, and $\hat{t}_{i,j}$ is the predicted probability of generating the $i$-th word at time step $j$.

During the validation and test phases, the Chg2Cap model generates captions for the input image pairs using an autoregressive approach. At the beginning of the decoding process, the word embedding is initialized with the ``START" token. The decoder then predicts the next word in the sentence by focusing on the relevant image representations and previously generated words, and appends this word to the end of the sentence. This iterative process continues until the model generates the ``END" token, indicating that the caption is complete.
\section{Experiments and Results}
In order to conduct a comprehensive evaluation of the proposed Chg2Cap method, we compared it with existing remote sensing change captioning methods as well as natural image change captioning methods on two benchmark remote sensing change captioning (RSCC) datasets. The detailed comparison and evaluation results are presented in this section.
\subsection{Datasets}
\subsubsection{Dubai-CC Dataset}
The Dubai-CC dataset, created by Hoxha et al. \cite{hoxha2022change}, provides a comprehensive description of the urbanization changes in the Dubai area. The dataset comprises image pairs acquired by the Enhanced Thematic Mapper Plus (ETM+) sensor onboard Landsat 7, captured on May 19, 2000, and June 16, 2010. To ensure accurate identification and description of the changes, the original images are cropped into 500 tiles of sizes 50 $\times$ 50, with five change descriptions annotated for each small bitemporal tile, referencing Google Maps and publicly available documents. The dataset includes 2,500 independent descriptions, with a maximum length of 23 words and an average length of 7.35 words. We adopted the default experimental settings outlined in \cite{hoxha2022change}. The dataset was split into training, validation, and testing sets, comprising 300, 50, and 150 bitemporal tiles, respectively. Our vocabulary is derived from all words present in the annotated sentences, resulting in 262 unique words. The experiments involved upsampling the cropped images to a size of 256 $\times$ 256 before inputting them into the change captioning networks.

\subsubsection{LEVIR-CC Dataset}
The LEVIR-CC dataset is derived from a building change detection dataset comprising 637 very high-resolution (0.5 m/pixel) bitemporal images of size 1,024 $\times$ 1,024, acquired over 20 regions of Texas, USA \cite{chen2020spatial}. The image pairs, with a time span of 5-14 years, were obtained from the Google Earth API. To facilitate its use in RSCC, the LEVIR-CC dataset was constructed by exploiting 10,077 small bitemporal tiles of size 256 $\times$ 256 pixels, with each tile annotated as containing changes or no changes. The dataset consists of 5038 image pairs with changes and 5039 without changes, with each image pair having five different sentence descriptions describing the nature of changes between the two acquisitions. The maximum sentence length is 39 words, with an average of 7.99 words.  As introduced in \cite{liu2022remote}, we followed the detailed description and employed the default splitting that resulted in 6815, 1333, and 1929 image pairs for training, validation, and testing sets, respectively. Our vocabulary includes only those words that appear at least five times in the annotated sentences, resulting in 497 unique words.

\subsection{Experimental Setup}

\subsubsection{Evaluation Metrics} 
The evaluation of the captioning model depends on the extent to which the generated descriptive sentences conform to human judgments about differences between bitemporal images. Automatic evaluation metrics can quantify the accuracy of the generated sentences based on annotated reference sentences. In this work, we utilized four commonly used metrics in both image captioning \cite{huang2020image, stefanini2022show, ji2022knowing} and change captioning \cite{qiu20203d,hosseinzadeh2021image,ak2022learning} tasks to evaluate the accuracy of all methods: BLEU-N (where N=1,2,3,4) \cite{papineni2002bleu}, ROUGE-L \cite{lin2004rouge}, METEOR \cite{banerjee2005meteor}, and CIDEr-D \cite{vedantam2015cider}. These metrics calculate the consistency between the predicted sentences and the references; thus, higher metric scores indicate a more significant similarity between the generated and reference sentences and higher captioning accuracy.

\subsubsection{Implementation Details}
The deep learning methods presented in this study are implemented using the PyTorch framework and trained and evaluated on an NVIDIA A100 graphics processing unit. During training, the Adam optimizer \cite{kingma2014adam} is employed with an initial learning rate of 0.0001 and a weight decay of 0.5 after 5 epochs. The training process is terminated after 50 epochs and the training batch size is set to 32. After each epoch, the model is evaluated on the validation set, and the best-performing model, based on the highest BLEU-4 score, is selected for evaluation on the test set.

\begin{table}
\centering
\caption{The performance of the model in different depths on the Dubai-CC dataset}
\resizebox{0.49\textwidth}{!}{
\begin{threeparttable}
\begin{tabular}{cc|ccccccc}
\toprule
\toprule[0.5pt]
D.H. & D.T. & BLEU-1   &  BLEU-2  & BLEU-3  & BLEU-4 & METEOR  & ROUGE-L   & CIDEr-D  \\ \midrule
1 & 1  & \textbf{79.93}   & \textbf{69.18}   & 59.36 & 50.57 & 31.63 & 61.50 & 107.72 \\ 
1 & 2  & 70.92   & 60.53   & 52.26 & 44.52 & 28.63 & 55.53 & 85.78 \\ 
1 & 3  & 66.98   & 55.82   & 48.54 & 40.97 & 26.00 & 51.33 & 83.89 \\ 
1 & 4  & 74.60   & 63.22   & 53.84 & 44.18 & 32.23 & 66.27 & 109.55 \\\midrule
2 & 1  & 74.50   & 66.91  & \textbf{60.70} & \textbf{54.72} & 33.01 & 62.83 & 108.73 \\ 
2 & 2  & 70.63  & 61.04   & 54.77 & 49.79 & 30.17 & 58.69 & 98.79 \\ 
2 & 3  & 64.90   & 56.70  & 50.78 & 45.99 & 27.14 & 57.76 & 99.45 \\ 
2 & 4  & 63.88   & 52.28   & 43.64 & 36.51 & 23.41 & 47.54 & 72.89 \\\midrule
3 & 1 & 79.21   & 68.77   & 59.34 & 49.56 & \textbf{34.70} & \textbf{67.37} & \textbf{126.22} \\ 
3 & 2  & 70.91   & 61.53  & 54.82 & 48.67 & 28.89 & 55.93 & 85.54 \\ 
3 & 3 & 76.41   & 65.33   & 57.01 & 48.29 & 31.76 & 62.39 & 117.67  \\ 
3 & 4  & 68.98   & 57.49   & 49.19 & 41.97 & 27.60 & 50.73 & 70.65 \\\midrule
4 & 1  & 76.42   & 67.49   & 60.25 & 52.84 & 33.76 & 62.32 & 117.95 \\
4 & 2  & 71.69   & 62.33   & 56.41 & 52.13 & 30.86 & 55.80 & 92.43 \\ 
4 & 3  & 70.13   & 60.73   & 53.62 & 46.95 & 28.62 & 58.12 & 90.15 \\
4 & 4  & 72.16   & 62.24  & 53.88 & 46.77 & 31.77 & 60.67 & 100.26 \\
\bottomrule[0.5pt] \bottomrule
\end{tabular}
\label{tab1}
\begin{tablenotes}
\footnotesize
\item[*] Note: D.H. denotes the depth of the HSA block in the attentive encoder, and D.T. denotes the depth of the transformer in the caption decoder. All the scores are reported in $\%$, and the best results are highlighted in \textbf{bold}. 
\end{tablenotes}
\end{threeparttable}}
\end{table}

To extract deep features from the input bitemporal image pairs, we use the ResNet-101 architecture that has been pre-trained on the ImageNet dataset, and the final fully convolutional layer is removed. Thus, the output of the feature extractor has 2048 channels. To ensure fair comparisons, the comparative methods also employ the same backbone settings. The number of heads in the multi-head attention mechanism is set to 8. For the parameters of the feed-forward layer in the attentive encoder, we set the channel of the linear layer to 512. The dimension of the word/image embeddings is set to 2048.

\begin{table}[t]
\centering
\caption{The performance of the model in different depths on the LEVIR-CC dataset}
\resizebox{0.49\textwidth}{!}{
\begin{threeparttable}
\begin{tabular}{cc|ccccccc}
\toprule
\toprule[0.5pt]
D.H. & D.T. & BLEU-1   &  BLEU-2  & BLEU-3  & BLEU-4 & METEOR  & ROUGE-L   & CIDEr-D  \\ \midrule
1 & 1  & 82.90 & 74.39 & 67.32 & 61.50 & 39.99 & 74.07 & 134.69 \\
1 & 2  & 82.03 & 73.36 & 66.26 & 60.47 & 38.32 & 71.97 & 129.47 \\ 
1 & 3  & 82.88 & 74.28 & 67.19 & 61.08 & 39.38 & 73.42 & 135.78 \\ 
1 & 4  & 82.85 & 74.33 & 67.40 & 61.65 & \textbf{40.08} & \textbf{74.45} & 136.02 \\\midrule
2 & 1  & 82.19 & 73.32 & 65.92 & 60.06 & 39.09 & 72.91 & 131.92 \\ 
2 & 2  & 81.15 & 72.19 & 64.64 & 58.51 & 38.33 & 72.28 & 128.12 \\ 
2 & 3  & 81.47 & 72.68 & 65.44 & 59.24 & 38.85 & 73.73 & 132.29 \\ 
2 & 4  & 81.71 & 72.57 & 65.24 & 59.30 & 39.17 & 73.08 & 134.17 \\\midrule
3 & 1  & \textbf{84.48} & \textbf{76.85} & \textbf{70.24} & \textbf{64.57} & 39.50 & \textbf{74.45} & \textbf{136.12} \\
3 & 2  & 81.96 & 73.31 & 65.94 & 59.77 & 38.71 & 72.49 & 129.92 \\ 
3 & 3  & 80.40 & 71.04 & 63.82 & 58.02 & 37.87 & 72.06 & 128.03 \\
3 & 4  & 79.78 & 71.67 & 64.97 & 59.42 & 37.42 & 71.00 & 123.07 \\\midrule
4 & 1  & 82.18 & 73.79 & 66.73 & 60.48 & 39.74 & 73.45 & 132.68 \\ 
4 & 2  & 80.70 & 71.66 & 64.48 & 58.42 & 38.51 & 72.17 & 130.17 \\ 
4 & 3  & 81.36 & 72.68 & 65.70 & 60.06 & 38.66 & 72.17 & 130.57 \\ 
4 & 4  & 81.40 & 72.71 & 65.72 & 59.82 & 38.20 & 72.93 & 130.64 \\
\bottomrule[0.5pt] \bottomrule
\end{tabular}
\label{tab2}
\begin{tablenotes}
\footnotesize
\item[*] Note: D.H. denotes the depth of the HSA block in the attentive encoder, and D.T. denotes the depth of the transformer in the caption decoder. All the scores are reported in $\%$, and the best results are highlighted in \textbf{bold}.
\end{tablenotes}
\end{threeparttable}}
\end{table}

\subsection{Parametric Analysis}

As previously described, the Chg2Cap model comprises an attentive encoder and caption decoder, each consisting of multiple layers of attention mechanisms. The depth of these networks can have a significant impact on the accuracy of generating descriptions for bitemporal images. In this section, we assess the performance of the proposed Chg2Cap model at various depths on both the Dubai-CC and LEVIR-CC datasets.
Tables \ref{tab1} and \ref{tab2} present the quantitative results on the validation set, where D.H. represents the depth of the hierarchical self-attention (HSA) blocks in the attentive encoder, and D.T. denotes the depth of the transformer decoder in the caption decoder. It is consistently observed that Chg2Cap achieves a more balanced performance when the transformer decoder has a shallower depth. This observation suggests that an excessively complex decoder may adversely influence the word generation performance during training. Furthermore, when the value of D.T. is fixed to 1, the proposed Chg2Cap model performs better when D.H. is set to 3. Through empirical analysis, we suggest that setting the values of D.H. and D.T. to 3 and 1, respectively, results in optimal performance for the Chg2Cap model.

\begin{table*}
\centering
\caption{Ablation study of different operations of the HSA block on the Dubai-CC dataset}
\begin{threeparttable}
\begin{tabular}{ccc|ccccccc|cc}
\toprule
\toprule[0.5pt]
Pos.Emb & DSA & JSA & BLEU-1   &  BLEU-2  & BLEU-3  & BLEU-4 & METEOR  & ROUGE-L   & CIDEr-D  & Parameters & Inference Time\\ \midrule
\checkmark & \checkmark & \checkmark & \textbf{72.04}   & \textbf{60.18}   & \textbf{50.84} & \textbf{41.70} & \textbf{28.92} & \textbf{58.66} & \textbf{92.49} & 285.5M & 6.99\\ 
\scalebox{0.75}{\usym{2613}} &\checkmark &\checkmark & 66.08   & 53.51   & 44.17 & 35.39 & 25.69 & 54.05 & 83.45 & 285.5M & 6.80\\\midrule
\checkmark & \scalebox{0.75}{\usym{2613}}  & \checkmark & 70.30 & 57.71   & 46.40 & 36.42 & 26.34 & 54.14 & 77.13 & 231.0M & 6.68\\
\checkmark & \checkmark & \scalebox{0.75}{\usym{2613}}  & 66.23   & 53.27   & 43.06 & 34.23 & 25.43 & 52.82 & 76.98 & 174.3M & 6.73\\
\scalebox{0.75}{\usym{2613}} & \scalebox{0.75}{\usym{2613}} &\scalebox{0.75}{\usym{2613}}& 67.76   & 56.39   & 47.32 & 39.00 & 27.86 & 57.38 & 88.95 & 115.5M & 6.37\\ 
\bottomrule[0.5pt] \bottomrule
\end{tabular}
\label{tab3}
\begin{tablenotes}
\footnotesize
\item[*] Note: Pos.Emb denotes the positional embedding initialization, DSA denotes the dual self-attention unit, and JSA denotes the joint self-attention unit. All the scores are reported in $\%$, and the best results are highlighted in \textbf{bold}. The inference time measures the total duration, in seconds, the network takes to process the entire testing set.
\end{tablenotes}
\end{threeparttable}
\end{table*}

\begin{table*}
\centering
\caption{Ablation study of different operations of the HSA block on the LEVIR-CC dataset}
\begin{threeparttable}
\begin{tabular}{ccc|ccccccc|cc}
\toprule
\toprule[0.5pt]
Pos.Emb & DSA & JSA & BLEU-1   &  BLEU-2  & BLEU-3  & BLEU-4 & METEOR  & ROUGE-L   & CIDEr-D  & Parameters & Inference Time\\ \midrule
\checkmark & \checkmark & \checkmark & \textbf{86.14} & \textbf{78.08} & \textbf{70.66} & \textbf{64.39} & \textbf{40.03} & \textbf{75.12} & \textbf{136.61} & 285.5M & 83.45\\ 
\scalebox{0.75}{\usym{2613}} &\checkmark &\checkmark & 83.24   & 74.20   & 66.69 & 60.70 & 39.58 & 73.81 & 133.54 & 285.5M & 82.93\\\midrule
\checkmark & \scalebox{0.75}{\usym{2613}}  & \checkmark & 82.90   & 73.63   & 65.84 & 59.33 & 38.15 & 72.26 & 130.19  & 231.0M & 75.82\\
\checkmark & \checkmark & \scalebox{0.75}{\usym{2613}}  & 84.31 & 75.63   & 68.28   & 62.28 & 39.57 & 74.62 & 135.61 & 174.3M & 79.78\\
\scalebox{0.75}{\usym{2613}} & \scalebox{0.75}{\usym{2613}} &\scalebox{0.75}{\usym{2613}}& 83.30 & 73.61   & 66.43   & 60.01 & 38.19 & 73.30 & 131.24 & 115.5M & 73.52\\ 
\bottomrule[0.5pt] \bottomrule
\end{tabular}
\label{tab4}
\begin{tablenotes}
\footnotesize
\item[*] Note: Pos.Emb denotes the positional embedding initialization, DSA denotes the dual self-attention unit, and JSA denotes the joint self-attention unit. All the scores are reported in $\%$, and the best results are highlighted in \textbf{bold}. The inference time measures the total duration, in seconds, the network takes to process the entire testing set.
\end{tablenotes}
\end{threeparttable}
\end{table*}

\begin{table*}
\centering
\caption{Ablation study of different operations of the ResBlock on the Dubai-CC dataset.}
\begin{threeparttable}
\begin{tabular}{cc|ccccccc|cc}
\toprule
\toprule[0.5pt]
cos & res & BLEU-1   &  BLEU-2  & BLEU-3  & BLEU-4 & METEOR  & ROUGE-L   & CIDEr-D  & Parameters & Inference Time \\ \midrule
\checkmark & \checkmark & \textbf{72.04}   & \textbf{60.18}   & \textbf{50.84} & \textbf{41.70} & \textbf{28.92} & 58.66 & 92.49 & 285.5M & 6.99\\ 
\scalebox{0.75}{\usym{2613}} & \checkmark  & 69.66   & 58.06   & 48.67 & 38.76 & 27.81 & \textbf{59.42} & \textbf{97.91} & 271.9M & 6.47\\ 
\checkmark & \scalebox{0.75}{\usym{2613}}  & 67.86   & 54.34   & 43.86 & 33.66 & 25.54 & 55.09 & 81.35 & 285.5M & 6.50\\ 
\scalebox{0.75}{\usym{2613}} & \scalebox{0.75}{\usym{2613}} & 64.55   & 51.91   & 42.85 & 35.29 & 24.15 & 51.44 & 76.30 & 271.9M & 6.33\\
\bottomrule[0.5pt] \bottomrule
\end{tabular}
\begin{tablenotes}
\footnotesize
\item[*] Note: cos denotes the cosine similarity mask, and res denotes the residual block. All the scores are reported in $\%$, and the best results are highlighted in \textbf{Bold}. The inference time measures the total duration, in seconds, the network takes to process the entire testing set.
\end{tablenotes}
\end{threeparttable}
\label{tab5}
\end{table*}

\begin{table*}
\centering
\caption{Ablation study of different operations of the ResBlock on the LEVIR-CC dataset.}
\begin{threeparttable}
\begin{tabular}{cc|ccccccc|cc}
\toprule
\toprule[0.5pt]
cos & res & BLEU-1   &  BLEU-2  & BLEU-3  & BLEU-4 & METEOR  & ROUGE-L   & CIDEr-D  & Parameters & Inference Time\\ \midrule
\checkmark & \checkmark & \textbf{86.14} & \textbf{78.08} & \textbf{70.66} & \textbf{64.39} & \textbf{40.03} & \textbf{75.12} & \textbf{136.61} & 285.5M & 83.45\\ 
\scalebox{0.75}{\usym{2613}} & \checkmark  & 81.55   & 72.28   & 64.58 & 58.32 & 38.77 & 72.78 & 130.76 & 271.9M  & 79.40\\ 
\checkmark & \scalebox{0.75}{\usym{2613}}  & 81.88   & 72.52   & 64.71 & 58.67 & 38.29 & 72.42 & 128.92 & 285.5M & 83.33\\ 
\scalebox{0.75}{\usym{2613}} & \scalebox{0.75}{\usym{2613}} & 80.54   & 70.88   & 63.36 & 57.33 & 37.59 & 71.82 & 126.54 & 271.9M & 78.08\\
\bottomrule[0.5pt] \bottomrule
\end{tabular}
\begin{tablenotes}
\footnotesize
\item[*] Note: cos denotes the cosine similarity mask, and res denotes the residual block. All the scores are reported in $\%$, and the best results are highlighted in \textbf{Bold}. The inference time measures the total duration, in seconds, the network takes to process the entire testing set.
\end{tablenotes}
\end{threeparttable}
\label{tab6}
\end{table*}

\subsection{Ablation Studies}
Here, we present the results of our ablation study, which aims to evaluate the effectiveness of the hierarchical self-attention (HSA) block and the residual block (ResBlock) in the attention decoder of the Chg2Cap model.
\subsubsection{Effects of the HSA Block}
 The HSA block is composed of a positional embedding process and a stack of hierarchical self-attention layers, which include a dual self-attention unit and a joint self-attention unit. We investigate the contributions of different components in the HSA block to the overall performance of the Chg2Cap network. The results of our experiments on the Dubai-CC and the LEVIR-CC datasets are presented in Tables \ref{tab3} and \ref{tab4}, respectively. In these tables,  Pos.Emb denotes the positional embedding initialization, DSA denotes the dual self-attention unit, and JSA denotes the joint self-attention unit. While the inclusion of the HSA block does introduce a slight increase in the network's inference time, the delay per frame remains remarkably low, measuring less than 5 milliseconds when the HSA block is not utilized. These results underscore the substantial enhancement in caption accuracy achieved by the HSA block across both datasets. Furthermore, our findings reveal that the optimal performance is achieved when all three operations are combined.

\subsubsection{Effects of the ResBlock} 
In this part, we investigate the efficacy of the ResBlock in the attentive encoder, which comprises a cosine mask calculation process and a residual block with three 2D convolutional layers using a ReLU activation function. To evaluate the impact of these two operations on the Chg2Cap model's performance, we conducted ablation experiments on both the Dubai-CC and the LEVIR-CC datasets, and the results are presented in Tables \ref{tab5} and \ref{tab6}, respectively. We denote the cosine mask factor as ``cos'' and the residual block as ``res''. When neither the cosine mask nor the residual block is included, the concatenation of the two retrieved features is directly passed to the caption decoder as the image embedding. Our experimental results consistently validate that computing the cosine similarity between the two features of the bitemporal image pair does not increase the network’s parameters, while improving the model’s ability to locate change-related objects. Additionally, the inclusion of the residual block contributes to the network’s performance. In comparison to the minor increase in inference time, our designed ResBlock effectively enhances the model's proficiency in change captioning.

\subsection{Qualitative Visualization}
\subsubsection{Caption Generation}
\begin{figure*}
    \centering
    \includegraphics[width=\linewidth]{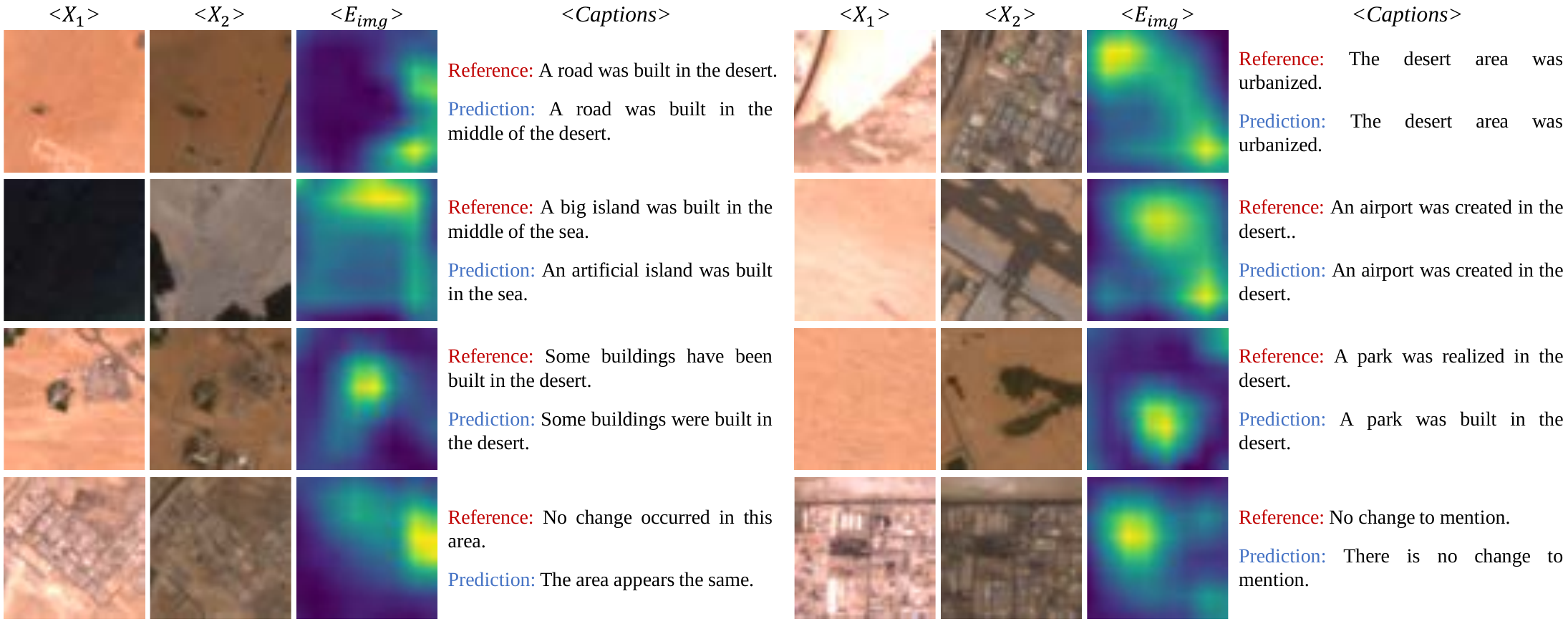}
    \caption{Visualized image embeddings and change captioning examples generated by Chg2Cap in the Dubai-CC dataset.}
    \label{fig5}
\end{figure*}

\begin{figure*}
    \centering
    \includegraphics[width=\linewidth]{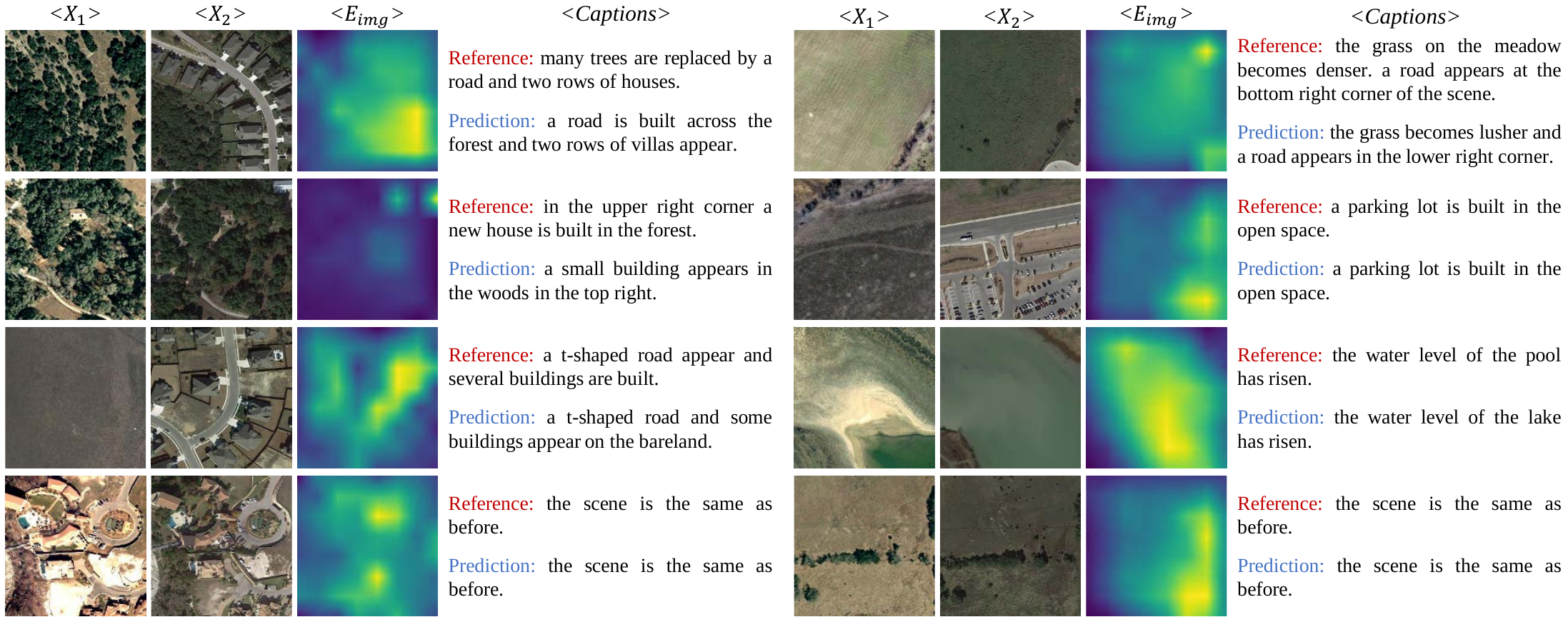}
    \caption{Visualized image embeddings and change captioning examples generated by Chg2Cap in the LEVIR-CC dataset.}
    \label{fig6}
\end{figure*}

To assess the quality of the change captions produced by our proposed Chg2Cap method, we conducted a qualitative evaluation by selecting several representative scenes from both the Dubai-CC and LEVIR-CC datasets. We visualized the image embeddings and the predicted captions generated by the caption decoder, as shown in Figs. \ref{fig5} and \ref{fig6}, where $X_1$ and $X_2$ represent the images captured at time 1 and time 2, respectively, and $E_{img}$ is the visualized image embedding extracted by the attentive encoder. Through visualizing image embedding features, we aim to capture change information and gain valuable insights into which parts of the scene are receiving more attention.


As noted earlier, the Dubai-CC dataset documents the urban development of Dubai, a city located in a vast desert area near the sea. With RGB bands spatial resolution of only 30 m, many typical land covers such as parks, buildings, roads, and islands occupy a small area in the image tiles, which may mislead the network when attempting to accurately describe the changes between the bitemporal images. However, it can still be observed from Fig. \ref{fig5} that our Chg2Cap method is capable of distinguishing changes and generating relatively realistic captions for each image pair, which appear reasonable when compared to human-annotated descriptions. Moreover, the visualized image embeddings indicate that the network can generally locate changes and highlight them as input to the decoder, potentially enabling the decoder to generate more accurate descriptions for the corresponding image pairs. The examples of unchanged scenes also demonstrate what the network focuses on when recognizing unchanged objects.

The change captioning results generated by the proposed Chg2Cap method for the LEVIR-CC dataset are presented in Fig. \ref{fig6}. The LEVIR-CC dataset has a higher spatial resolution than the Dubai-CC dataset, which results in more detailed land cover information, including vegetation, buildings, water areas, and roads. Chg2Cap demonstrates sensitivity to these land covers, accurately identifying changes and generating appropriate captions. The visualized image embeddings in Fig. \ref{fig6} support the ability of the attentive encoder to locate and highlight changes between the bitemporal images, potentially providing valuable referential image embeddings for the caption generator.

\subsubsection{Attention Weights}
To gain a deeper understanding of how the Chg2Cap model generates change captions, we conducted a further analysis of the attention weights of the transformer decoder. This allowed us to identify the specific areas within the image that the network pays attention to when generating the captions for different land covers.

\begin{figure}
    \centering
    \includegraphics[width=\linewidth]{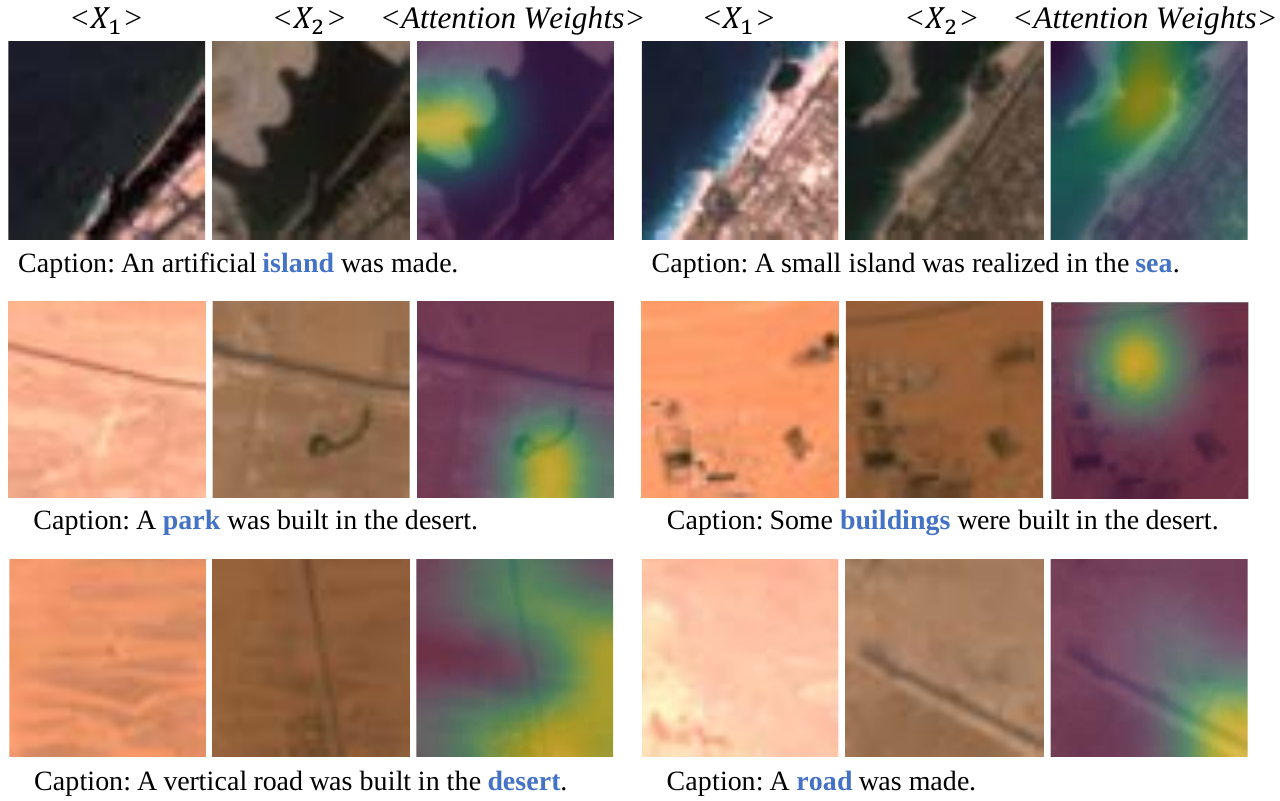}
    \caption{Visualized attention weights corresponding to specific land covers generated by Chg2Cap in the LEVIR-CC dataset. The corresponding words are marked in \textcolor{blue}{blue}, and the corresponding attention weights are superimposed on the image at time 2.}
    \label{fig7}
\end{figure}

\begin{figure}
    \centering
    \includegraphics[width=\linewidth]{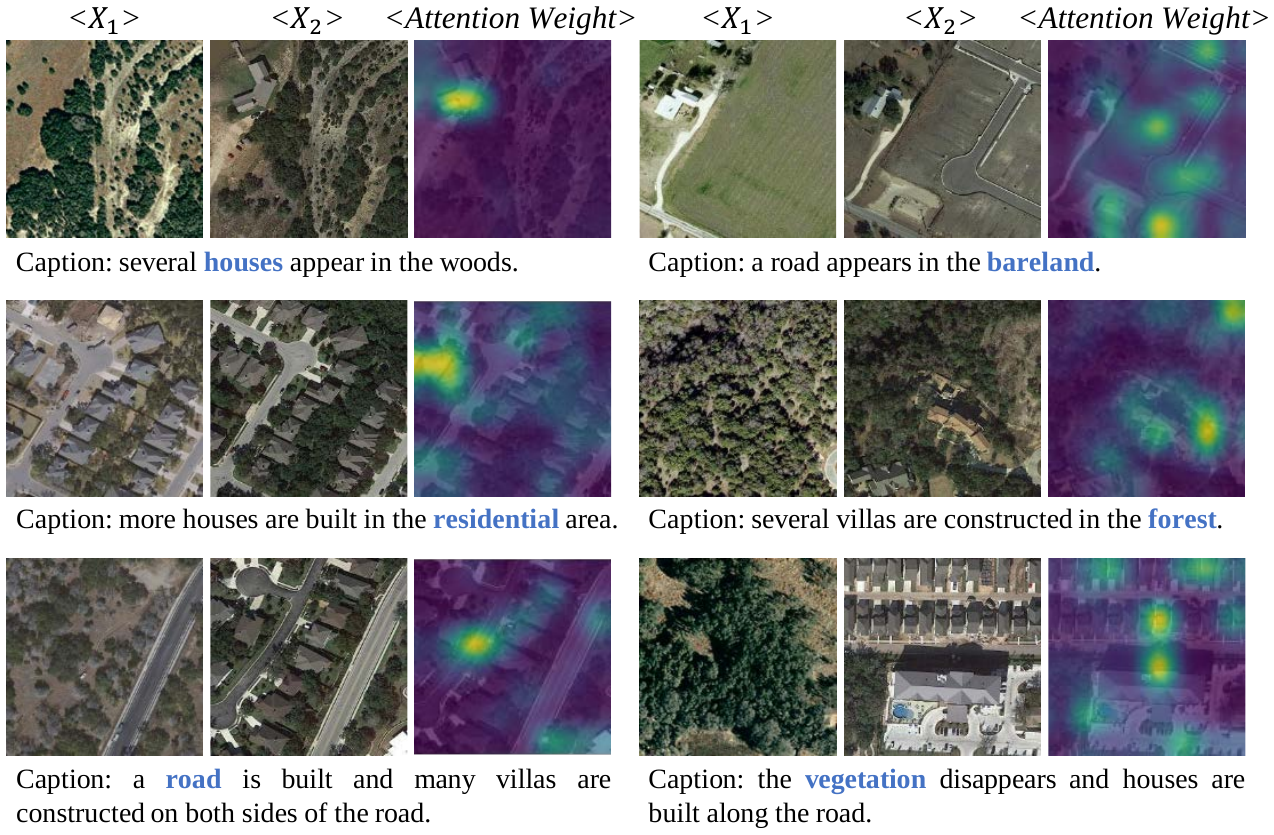}
    \caption{Visualized attention weights corresponding to specific land covers generated by Chg2Cap in the LEVIR-CC dataset. The corresponding words are marked in \textcolor{blue}{blue}, and the corresponding attention weights are superimposed on the image at time 2.}
    \label{fig8}
\end{figure}

We present the attention weights for some typical land covers in the Dubai-CC dataset in Fig. \ref{fig7}. For small targets such as islands, parks, buildings, and roads, the attention weights accurately identify their approximate locations within the image. Conversely, for large-scale land covers such as the sea and the desert, the attention weights highlight a relatively larger area within the image. These findings suggest that the Chg2Cap model can focus its attention on different spatial scales depending on the size of the land cover in reality. In the LEVIR-CC dataset, we also visualized the attention weights of the Chg2Cap model and found that it can focus well on the corresponding change regions when generating object-related words. Furthermore, for words that represent a wide range of land covers, such as residential, bareland, forest, and vegetation, the model's attention is more widely spread across the overall image scene, as shown in Fig. \ref{fig8}. These results indicate that the Chg2Cap model can identify and attend to the most salient areas of the image effectively, to generate accurate and relevant captions.


\subsection{Comparison to the State of the Art}
We benchmark several state-of-the-art change captioning methods on the Dubai-CC and LEVIR-CC datasets, including three natural image change captioning methods, DUDA, MCCFormers-S, and MCCFormers-D, and three remote sensing image change captioning methods, CC+RNN, CC+SVM, and RSICCformer. A more detailed introduction to these methods follows.

\begin{itemize}
    \item DUDA \cite{park2019robust} proposes a dual dynamic attention model that aims to differentiate distractors and locate changes in ``before" and ``after" images. The authors employ a dynamic speaker based on Long Short-Term Memory (LSTM) \cite{hochreiter1997long} to describe semantic changes over the visual representations. The dynamic speaker uses a dynamic attention mechanism to transfer the visual features from the encoder as attention weights and then generates the predicted sentence based on the word embedding and the dynamically attended feature derived from attention weights and visual representations via the caption model.
    \item MCCFormers-S \cite{qiu2021describing} is a transformer-based change captioning method that extracts ``before" and ``after" feature maps from two images using ResNet-101. The two feature maps are linearly transformed and positionally encoded before being flattened into feature sequences and concatenated as one input to the transformer encoder and decoder.  
    \item MCCFormers-D \cite{qiu2021describing} adopts a Siamese cross-attention architecture to extract change-related objects. Similar to MCCFormers-S, the two feature maps extracted by the feature extractor are linearly transformed and positionally encoded before being flattened into feature sequences. The cross-attention mechanism \cite{lee2018stacked} is employed to capture relationships between the two feature maps. The dual feature maps are then fed into the transformer decoder to generate the change caption that describes the differences between the two images.
    \item CC+RNN \cite{hoxha2022change} first proposes a change captioning system for remote sensing images. It introduces an early fusion strategy that concatenates the two input images directly and utilizes a pre-trained CNN backbone to extract features. The generated captions are produced by a classical RNN, specifically a GRU architecture, which serves as the caption generator.
    \item CC+SVM \cite{hoxha2022change} employs a late fusion strategy that fuses the extracted features from the two input images before decoding the feature maps using a network of multiclass SVMs. The caption generator uses a cascade of $k$ SVM linear multiclass classifiers in conjunction with a recurrent model to predict a new word based on the bitemporal image features and the previously predicted word \cite{hoxha2021novel}.
    \item RSICCformer \cite{liu2022remote} proposes a transformer-based architecture for remote sensing image change captioning. The model consists of an image encoder, a dual-branch transformer encoder, and a text generator. The dual-branch transformer comprises a stack of cross-encoding and bitemporal feature fusion modules that capture the relationships between image features and fuse them. The transformer decoder is employed to generate the next token from the input image features and the preceding token in the sequence.
\end{itemize}

\begin{table*}
\centering
\caption{Quantitative Comparison Results on the Dubai-CC Datasets using ResNet-101 Architecture}
\begin{threeparttable}
\begin{tabular}{c|ccccccc}
\toprule
\toprule[0.5pt]
\textbf{Method} & BLEU-1   &  BLEU-2  & BLEU-3  & BLEU-4 & METEOR  & ROUGE-L   & CIDEr-D  \\ \midrule
DUDA\cite{park2019robust}     & 58.82   & 43.59   & 33.63 & 25.39 & 22.05  & 48.34 & 62.78 \\ 
MCCFormers-S\cite{qiu2021describing}   & 52.97   & 37.02   & 27.62& 22.57& 18.64 & 43.29 &53.81  \\ 
MCCFormers-D\cite{qiu2021describing}   & 64.65   & 50.45   & 39.36 &29.48& 25.09& 51.27 & 66.51  \\  
RSICCformer\cite{liu2022remote}     & 67.92  & 53.61  &41.37& 31.28  & 25.41 & 51.96 & 66.54 \\ 
\midrule
Baseline     &  65.75 & 51.87 & 42.01 & 33.09 & 25.17 & 52.06 & 72.23  \\
Ours         & \textbf{72.04}   & \textbf{60.18}   & \textbf{50.84} & \textbf{41.70} & \textbf{28.92} & \textbf{58.66} & \textbf{92.49}  \\ \bottomrule[0.5pt] \bottomrule
\end{tabular}
\label{tab7}
\begin{tablenotes}
\footnotesize
\item[*] Note: All the scores are reported in $\%$, and the best results are highlighted in \textbf{bold}.
\end{tablenotes}
\end{threeparttable}
\end{table*}

\begin{table*}
\centering
\caption{Quantitative Comparison Results on the Dubai-CC Dataset with CC+RNN and CC+SVM}
\begin{threeparttable}
\begin{tabular}{c|ccccccc}
\toprule
\toprule[0.5pt]
\textbf{Method} & BLEU-1   &  BLEU-2  & BLEU-3  & BLEU-4 & METEOR  & ROUGE-L   & CIDEr-D  \\ \midrule
CC-RNN-Sub\cite{hoxha2022change}        & 67.19  & 52.23    & 40.00 & 28.54 & 25.51 & 51.78 & 69.73 \\
CC-SVM-Sub\cite{hoxha2022change}        & 70.71    & 57.58    & 46.10 &35.50& 27.60 & 56.61 & 82.96 \\ 
\midrule
Baseline     &  66.20 & 50.77 & 43.54 & 34.39 & 26.02 & 55.23 & 71.26  \\
Ours         & \textbf{71.17}   & \textbf{58.46}   & \textbf{46.29} & \textbf{40.13} & \textbf{30.48} & \textbf{58.46} & \textbf{86.81}  \\ \bottomrule[0.5pt] \bottomrule
\end{tabular}
\label{tab8}
\begin{tablenotes}
\footnotesize
\item[*] Note: All methods utilize VGG-16 as the backbone. All the scores are reported in $\%$, and the best results are highlighted in \textbf{bold}.
\end{tablenotes}
\end{threeparttable}
\end{table*}

\begin{table*}
\centering
\caption{Quantitative Comparison Results on the LEVIR-CC Dataset using ResNet-101 Architecture}
\begin{threeparttable}
\begin{tabular}{c|ccccccc}
\toprule
\toprule[0.5pt]
\textbf{Method} & BLEU-1   &  BLEU-2  & BLEU-3  & BLEU-4 & METEOR  & ROUGE-L   & CIDEr-D  \\ \midrule
DUDA\cite{park2019robust}     & 81.44   & 72.22   & 64.24 &57.79 &37.15  & 71.04 & 124.32 \\ 
MCCFormers-S\cite{qiu2021describing}  & 82.16   & 72.95& 65.42& 59.41 & 38.26 & 72.10 &128.34  \\ 
MCCFormers-D\cite{qiu2021describing}   & 80.49   & 71.11   & 63.52 &57.34& 38.23& 71.40 & 126.85  \\ 
RSICCformer\cite{liu2022remote}     & 84.11  & 75.40  & 68.01& 61.93  & 38.79& 73.02 & 131.40 \\ 
\midrule
Baseline     & 82.41 & 73.10 & 65.29 & 59.02 & 38.71 & 72.47 & 130.88  \\
Ours         & \textbf{86.14} & \textbf{78.08} & \textbf{70.66} & \textbf{64.39} & \textbf{40.03} & \textbf{75.12} & \textbf{136.61}  \\ \bottomrule[0.5pt] \bottomrule
\end{tabular}
\label{tab9}
\begin{tablenotes}
\footnotesize
\item[*] Note: All the scores are reported in percentage (\%), and the best results are highlighted in \textbf{bold}.
\end{tablenotes}
\end{threeparttable}
\end{table*}
The performance evaluation of the proposed Chg2Cap model, along with three natural image change captioning methods and three remote sensing change captioning methods, is presented in Tables \ref{tab7} and \ref{tab8} using the Dubai-CC dataset. For a fair comparison, we separately compare the proposed Chg2Cap model with CC+RNN and CC+SVM methods, which also utilize VGG-16 as the backbone. The results demonstrate that the Chg2Cap model outperforms all the compared methods across all evaluation metrics. As previously mentioned, the Dubai-CC dataset has a spatial resolution of 30 m, which poses challenges in distinguishing many objects accurately. Additionally, the human-annotated sentences in this dataset have capitalized first letters, further complicating the accurate description of changed land covers. As a result, the accuracies of the comparative methods are relatively low, with all methods achieving BLEU-4 scores lower than 36\%. In contrast, our Chg2Cap model exhibits superior performance compared to the best-performing CC-SVM method, surpassing it by 4.6\% on BLEU-4 and 3.9\% on CIDEr-D. These results indicate that the captions generated by Chg2Cap are more similar to the reference sentences, and the relationships between the two images are better captured by the Chg2Cap model.

Additionally, Table \ref{tab9} presents the performance of the proposed Chg2Cap model compared with DUDA, MCCFormers-S, MCCFormers-D, and RSICCformer on the LEVIR-CC dataset. Notably, the LEVIR-CC dataset, with its higher spatial resolution, challenges all methods to achieve improved quantitative results. Despite this, the Chg2Cap method still outperforms other methods, surpassing the best-performing RSICCformer method by 5.3\% on BLEU-4 and 5.2\% on CIDEr-D.

Furthermore, it is worth noting that our baseline model does not incorporate the attentive encoder, which concatenates the two feature maps extracted by the feature extractor directly and feeds them as the image embedding into the caption decoder. The results obtained on both datasets demonstrate that the attentive encoder can significantly enhance the caption generation accuracy.

\section{Conclusion}
In this study, we proposed a novel attentive network for remote sensing change captioning, called Chg2Cap for short, which utilizes the power of transformer models in natural language processing. The proposed Chg2Cap is constructed by a CNN-based feature extractor that uses the pre-trained ResNet-101 as the backbone, an attentive encoder that consists of a hierarchical self-attention block and a ResBlock, and a caption decoder. The proposed method was evaluated on two public remote sensing datasets, Dubai-CC and LEVIR-CC, and achieved superior performance compared to both the remote sensing change captioning methods and the natural image change captioning methods. Through extensive experiments and analyses, we demonstrated the effectiveness of our proposed method in capturing detailed changes in various types of land covers, including buildings, vegetation, water areas, desert, and roads. The visualized captioning results and attention weights also provide insights into how the model generates accurate descriptions by highlighting the regions of interest in the images. Our work not only contributes to the field of remote sensing image analysis but also has the potential to benefit a range of applications by providing valuable information about changes, including urban planning, disaster management, and human-computer intelligent interaction in remote sensing. We believe that our proposed method can be further extended and optimized in the future to address more complex change captioning tasks and achieve even better performance.

\section*{Acknowledgments}
The authors would like to thank the authors of the comparative methods for sharing their codes, the contributors to the LEVIR-CC and Dubai-CC datasets, and the Institute of Advanced Research in Artificial Intelligence (IARAI) for its support.

\bibliographystyle{IEEEtran}
\bibliography{ref}

\vfill

\end{document}